\documentclass[11pt,a4paper]{article}
\pdfoutput=1
\usepackage[utf8]{inputenc}
\usepackage{microtype}
\usepackage{epsfig}
\usepackage{amssymb}
\usepackage{natbib}
\usepackage{graphicx}
\usepackage[colorlinks=false,allbordercolors={1 1 1}]{hyperref}
\usepackage{authblk}

\usepackage{amssymb}
\usepackage{pifont}
\newcommand{\cmark}{\ding{51}}%
\newcommand{\xmark}{\ding{55}}%

\bibpunct{(}{)}{;}{a}{,}{,}

\usepackage{xcolor}
\definecolor{maroon}{cmyk}{0, 0.87, 0.68, 0.32}
\definecolor{halfgray}{gray}{0.55}
\definecolor{ipython_frame}{RGB}{207, 207, 207}
\definecolor{ipython_bg}{RGB}{247, 247, 247}
\definecolor{ipython_red}{RGB}{186, 33, 33}
\definecolor{ipython_green}{RGB}{0, 128, 0}
\definecolor{ipython_cyan}{RGB}{64, 128, 128}
\definecolor{ipython_purple}{RGB}{170, 34, 255}

\usepackage{listings}
\lstdefinelanguage{iPython}{
    morekeywords={access,and,break,class,continue,def,del,elif,else,except,exec,finally,for,from,global,if,import,in,is,lambda,not,or,pass,print,raise,return,try,while},%
    morekeywords=[2]{abs,all,any,basestring,bin,bool,bytearray,callable,chr,classmethod,cmp,compile,complex,delattr,dict,dir,divmod,enumerate,eval,execfile,file,filter,float,format,frozenset,getattr,globals,hasattr,hash,help,hex,id,input,int,isinstance,issubclass,iter,len,list,locals,long,map,max,memoryview,min,next,object,oct,open,ord,pow,property,range,raw_input,reduce,reload,repr,reversed,round,set,setattr,slice,sorted,staticmethod,str,sum,super,tuple,type,unichr,unicode,vars,xrange,zip,apply,buffer,coerce,intern},%
    sensitive=true,%
    morecomment=[l]\#,%
    morestring=[b]',%
    morestring=[b]",%
    morestring=[s]{'''}{'''},
    morestring=[s]{"""}{"""},
    morestring=[s]{r'}{'},
    morestring=[s]{r"}{"},%
    morestring=[s]{r'''}{'''},%
    morestring=[s]{r"""}{"""},%
    morestring=[s]{u'}{'},
    morestring=[s]{u"}{"},%
    morestring=[s]{u'''}{'''},%
    morestring=[s]{u"""}{"""},%
    literate=
    {á}{{\'a}}1 {é}{{\'e}}1 {í}{{\'i}}1 {ó}{{\'o}}1 {ú}{{\'u}}1
    {Á}{{\'A}}1 {É}{{\'E}}1 {Í}{{\'I}}1 {Ó}{{\'O}}1 {Ú}{{\'U}}1
    {à}{{\`a}}1 {è}{{\`e}}1 {ì}{{\`i}}1 {ò}{{\`o}}1 {ù}{{\`u}}1
    {À}{{\`A}}1 {È}{{\'E}}1 {Ì}{{\`I}}1 {Ò}{{\`O}}1 {Ù}{{\`U}}1
    {ä}{{\"a}}1 {ë}{{\"e}}1 {ï}{{\"i}}1 {ö}{{\"o}}1 {ü}{{\"u}}1
    {Ä}{{\"A}}1 {Ë}{{\"E}}1 {Ï}{{\"I}}1 {Ö}{{\"O}}1 {Ü}{{\"U}}1
    {â}{{\^a}}1 {ê}{{\^e}}1 {î}{{\^i}}1 {ô}{{\^o}}1 {û}{{\^u}}1
    {Â}{{\^A}}1 {Ê}{{\^E}}1 {Î}{{\^I}}1 {Ô}{{\^O}}1 {Û}{{\^U}}1
    {œ}{{\oe}}1 {Œ}{{\OE}}1 {æ}{{\ae}}1 {Æ}{{\AE}}1 {ß}{{\ss}}1
    {ç}{{\c c}}1 {Ç}{{\c C}}1 {ø}{{\o}}1 {å}{{\r a}}1 {Å}{{\r A}}1
    {€}{{\EUR}}1 {£}{{\pounds}}1
    {^}{{{\color{ipython_purple}\^{}}}}1
    {=}{{{\color{ipython_purple}=}}}1
    {+}{{{\color{ipython_purple}+}}}1
    {*}{{{\color{ipython_purple}$^\ast$}}}1
    {/}{{{\color{ipython_purple}/}}}1
    {+=}{{{+=}}}1
    {-=}{{{-=}}}1
    {*=}{{{$^\ast$=}}}1
    {/=}{{{/=}}}1,
    literate=
    *{-}{{{\color{ipython_purple}-}}}1
     {?}{{{\color{ipython_purple}?}}}1,
    identifierstyle=\color{black}\ttfamily,
    commentstyle=\color{ipython_cyan}\ttfamily,
    stringstyle=\color{ipython_red}\ttfamily,
    keepspaces=true,
    showspaces=false,
    showstringspaces=false,
    basicstyle=\scriptsize,
    keywordstyle=\color{ipython_green}\ttfamily,
}

\begin{document}

\title{ELFI: Engine for Likelihood-Free Inference}

\author[1]{Jarno Lintusaari}
\author[1]{Henri Vuollekoski}
\author[1]{Antti Kangasrääsiö}
\author[1]{Kusti Skytén}
\author[1]{Marko Järvenpää}
\author[1]{Pekka Marttinen}
\author[2]{Michael U. Gutmann}
\author[1*]{Aki Vehtari}
\author[3*]{Jukka Corander}
\author[1*]{Samuel Kaski}

\affil[1]{Department of Computer Science, Aalto University, Helsinki, Finland}
\affil[2]{School of Informatics, The University of Edinburgh, Edinburgh, United Kingdom}
\affil[3]{Department of Biostatistics, University of Oslo, Oslo, Norway}
\affil[*]{equal contribution}
\affil[ ]{\textit {\{jarno.lintusaari,henri.vuollekoski,antti.kangasraasio,kusti.skyten, marko.j.jarvenpaa,pekka.marttinen,aki.vehtari,samuel.kaski\}@aalto.fi, michael.gutmann@ed.ac.uk, jukka.corander@medisin.uio.no}}

\maketitle

\begin{abstract}
Engine for Likelihood-Free Inference (ELFI) is a Python software library for performing likelihood-free inference (LFI).
ELFI provides a convenient syntax for arranging components in LFI, such as priors, simulators, summaries or distances, to a network called ELFI graph.
The components can be implemented in a wide variety of languages.
The stand-alone ELFI graph can be used with any of the available inference methods without modifications.
A central method implemented in ELFI is Bayesian Optimization for Likelihood-Free Inference (BOLFI), which has recently been shown to accelerate likelihood-free inference up to several orders of magnitude by surrogate-modelling the distance.
ELFI also has an inbuilt support for output data storing for reuse and analysis, and supports parallelization of computation from multiple cores up to a cluster environment.
ELFI is designed to be extensible and provides interfaces for widening its functionality.
This makes the adding of new inference methods to ELFI straightforward and automatically compatible with the inbuilt features.

\end{abstract}
\pagebreak

\section{Introduction}

Engine for Likelihood-Free Inference (ELFI) is a statistical software package for likelihood-free inference written in Python.
The term ``likelihood-free inference" (LFI) refers to a family of inference methods that can be used when the likelihood function is not computable or otherwise available, but it is possible to simulate from the model \citep[see e.g.][]{Lintusaari2017}. Other names for likelihood-free inference or closely related approaches include Approximate Bayesian Computation (ABC) \citep[see e.g.][]{Marin2012, Lintusaari2017}, simulator-based inference, approximative Bayesian inference and indirect inference.

In LFI, generative models are commonly composed of priors and user-specified simulators. The inference is based on the outputs of the generative model,
that is, on the simulated data for various parameter configurations, as opposed to the likelihoods of the observed data under the configurations.
To facilitate the inference, the observed and simulated data are usually summarized after which distances between the summaries are taken.
In ELFI, the simulators, summaries, distances, etc. are called components and can be implemented in a wide variety of languages.

One of the main features in ELFI is the convenient syntax of combining all of the components into a single network (Figure \ref{fig:example}) that we call an ELFI graph. Once the ELFI graph is specified, it can be used with any of the available inference algorithms. ELFI also supports parallelization of the inference from a single computer up to a computational cluster, and storing the generated data for reuse, post-processing and further analysis.
ELFI has emerged from the prior research on the subject by the authors \citep{Lintusaari2016, Gutmann2016, Lintusaari2017, Kangasraasio2017inferring} and was used by \citet{Kangasraasio2017inferring} and \citet{kangasraasio2017inverse}.

\section{Software Design Principles}
ELFI is designed to support likelihood-free inference research both from the practitioners' and methodologists' point of view.
We aim for an easy-to-use ecosystem where practitioners will find the state-of-the-art inference methods, whereas methodologists will find simulators along with accompanying ELFI graphs and data to aid in method development and assessment.

\subsection[Practitioners]{Features for Practitioners}

\begin{figure}
    \begin{minipage}[b]{.8\linewidth}
\begin{lstlisting}[language=iPython]
# Define the simulator, the summary and the observed data
def simulator(t1, t2, batch_size=1, random_state=None):
    # Implementation comes here. Return `batch_size` 
    # simulations wrapped to a NumPy array.
def summary(data, argument=0):
    # Implementation comes here...
y = # Observed data, as one element of a batch.

# Specify the ELFI graph
t1 = elfi.Prior('uniform', -2, 4)
t2 = elfi.Prior('normal', t1, 5)  # depends on t1
SIM = elfi.Simulator(simulator, t1, t2, observed=y)
S1 = elfi.Summary(summary, SIM)
S2 = elfi.Summary(summary, SIM, 2)
d = elfi.Distance('euclidean', S1, S2)

# Run the rejection sampler
rej = elfi.Rejection(d, batch_size=10000)
result = rej.sample(1000, threshold=0.1)
\end{lstlisting}
    \end{minipage}
    \begin{minipage}[b]{.19\linewidth}
            \includegraphics[width=1.8cm, trim={1cm 1cm 2cm 1cm}]{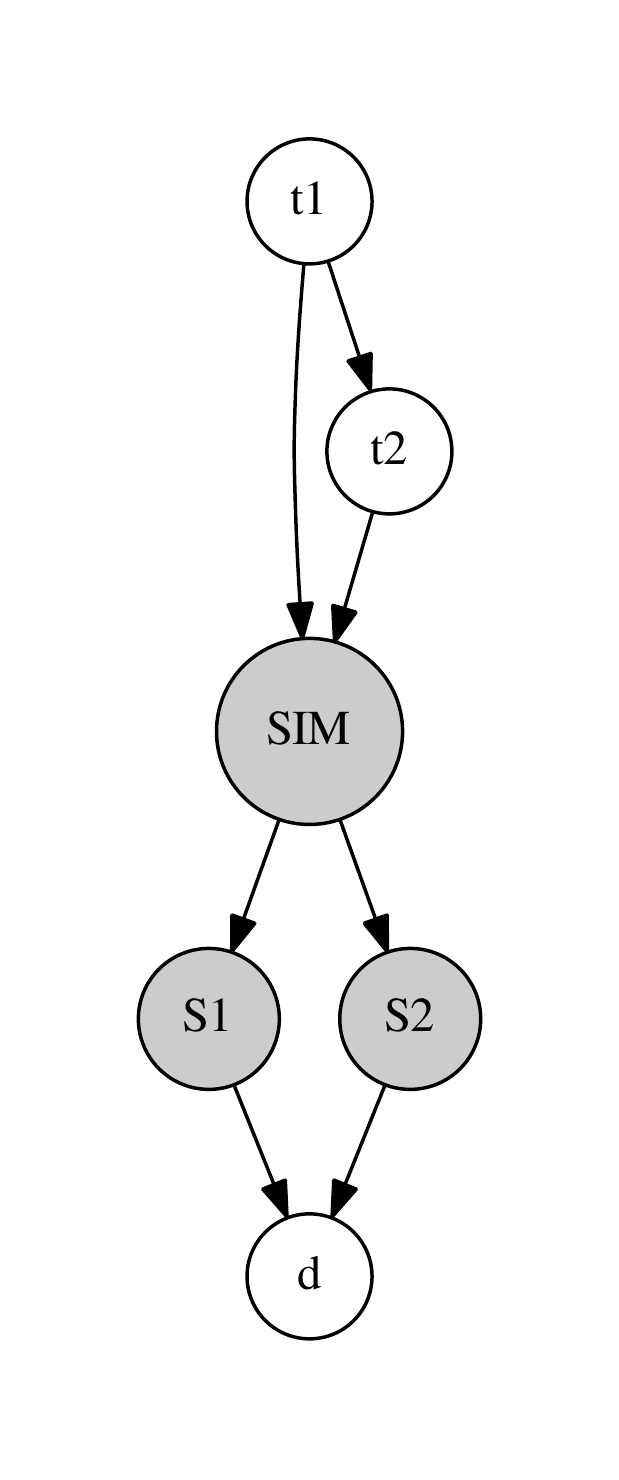}
    \end{minipage}
    \caption{Example of an ELFI graph and of running ABC rejection sampling in ELFI. The observed data are given to the operation that produces corresponding output. Two summaries are defined, where summary S2 is given an additional argument 2.}
	\label{fig:example}
\end{figure}

For practitioners ELFI provides a convenient interface for quickly arranging the components needed in LFI into an ELFI graph. Inherently ELFI graphs are directed acyclic graphs (DAGs), that represent how quantities used by the inference algorithm (e.g. distances) are computed (Figure \ref{fig:example}). The DAG structure
makes it possible to construct detailed hierarchies between the components (nodes). Under the hood, the ELFI graph is converted to a computation graph that will include e.g. nodes for the observed data (see the documentation\footnotemark[1] for more details). The nodes (components) are either data or operations that output data. Users are free to implement components as needed, but ELFI provides also ready made implementations for common components. Once specified, the ELFI graph can be directly used with any of the available inference methods.
We have provided an initial set of methods that can handle different types of scenarios: basic rejection sampling for cheap simulators, the general-purpose sequential Monte Carlo as well as BOLFI \citep{Gutmann2016} for expensive simulators. BOLFI combines probabilistic modelling of the distance with decision-making under uncertainty to decide for which parameter value to next run the simulator, significantly reducing the number of simulations needed.

Since likelihood-free inference often requires a moderate amount of experimentation (e.g. trying different summary statistics) it is important that specifying the components is made flexible and that already generated data can be reused.
We found the DAG structure to be ideal for these tasks.
First, ELFI allows any part of the ELFI graphs (e.g. nodes and their dependencies) to be redefined keeping the rest of the structure intact.
Second, ELFI provides automatic storing of the full output of any node (e.g. the simulator).
These data will be automatically reused when for example the summaries are changed, potentially resulting in significant savings in compute time.
These features are demonstrated in the ELFI tutorial in the documentation.

Another important factor is the ability to use non-Python components in the ELFI graph.
For instance, it may not always be practical or even possible to rewrite existing simulators in Python.
ELFI provides both tools and examples in the documentation on how to use simulators written in other languages.

Other practical features include the ability to progress the inference iteratively and to stop early if necessary.
The provided visualization functions support assessing the current state of the inference.
Finally, the ELFI graph can be saved to a file and shared with others.
ELFI also guarantees that the results will be identical for the same seeds making the reproduction of the results easy.

\subsection[Methodologists]{Features for Methodologists}

For methodologists ELFI provides a convenient platform for testing new algorithms with models from the literature \citep[e.g.][]{Ricker54, Marin2012, Lintusaari2017} and comparing their performance against existing algorithms.
The framework provides means for parallelization, data storing, seeding of pseudo random number generation and other important technicalities out of the box.
The documentation includes instructions on how to implement new algorithms for ELFI.
One of the major benefits is that all existing ELFI graphs will be usable with the new algorithms without modifications.

\section{Performance and Scalability}

Performance is an important factor in computationally heavy inference such as LFI.
ELFI uses batches of computations to control execution performance and parallelization.
A batch consists of a fixed number of consecutive evaluations of a node in the ELFI graph before moving to the next (e.g. 100 draws from the prior and then 100 simulations using those parameters). The standard parallelization strategy is to compute multiple batches in parallel.
This provides several benefits.
First, the computation of a single batch can often be vectorized with, for example, NumPy \citep{numpy11} for many of the basic operations (e.g. computing summaries or distances), making them efficient in Python. This is especially beneficial when experimenting with different summaries and distances with precomputed simulations.
Batches are also often relatively constant in their time and memory consumption, allowing flexibility in planning the parallel execution of multiple batches.
This helps in avoiding unnecessary message passing, progressing the inference in meaningful steps, and makes it possible to know in advance the size of the returned output data for storing purposes.

\section{Comparison to Other Similar Software}

There exist multiple LFI libraries for parameter inference.
Many of them are either restricted to a specific problem domain \citep[][]{Liepe2014, Cornuet2014, carl}, or require existing simulated data \citep{Thornton2009, Csillery2012, Nunes2015}. Edward \citep{tran2016} provides some LFI methods with a GPU acceleration, but requires the simulator to be differentiable. ELFI makes no extra requirements for the simulator (or other components), and can also be used with implementations taking benefit of hardware accelerations (e.g. GPU). General-purpose LFI software similar to ELFI are, to our knowledge, ABCtoolbox \citep{Wegmann2010}, EasyABC \citep{Jabot2013}, and ABCpy \citep{Dutta2017}.
A relatively recent categorization of LFI software is provided by \citet{Nunes2015}.

Among the general-purpose LFI software, only ELFI separates the LFI component specification from inference (Table \ref{table:comparison}). The graph-based specification provides considerable flexibility in both defining the components and experimenting with them.
For example, it is possible to embed multiple simulators into a single ELFI graph without modifying their codes. We refer the reader to the documentation for illustrations.\footnote{ELFI documentation can be found at \url{http://elfi.readthedocs.io}.}

Regarding parallelization, EasyABC supports multiple cores on a single computer while the others can also run in cluster environments.
By default, ABCpy uses Spark \citep{Spark} and ELFI ipyparallel for parallelization but both can be used with alternative backends.\footnote{The ipyparallel project can be found at \url{https://github.com/ipython/ipyparallel}.} ABCtoolbox does not provide a parallel solution out of the box.

ABCtoolbox, EasyABC and ELFI support reusing generated data.
ELFI is more flexible in that it allows the output of any node of the ELFI graph to be stored, and it automatically uses that data to compute the output of its current or future child nodes.
There is thus no need to manually transform existing data.

Only ELFI supports advancing the inference sample-by-sample, which facilitates debugging and enables e.g. convergence monitoring and early stopping. Also, ELFI is currently the only general-purpose software to implement the BOLFI method \citep{Gutmann2016}, which can handle expensive-to-evaluate simulators outside the reach of other methods.

\begin{center}
\begin{table}
  \tiny
  \begin{tabular}{ l  l  l  l  l  l  l  l }
    Software & Language & Latest release & Data reuse & Parallelization & Graph-based & Iterative processing \\
    \hline
    ABCtoolbox & C++ & 2009 & Partial & cluster (manual) & \xmark & \xmark \\
    EasyABC & R & 2015 & Partial & local & \xmark & \xmark \\
    ABCpy & Python & 2017 & \xmark & local and cluster & \xmark & \xmark \\
    ELFI & Python & 2017 & \cmark & local and cluster & \cmark & \cmark \\
  \end{tabular}
  \caption{Comparison of general-purpose LFI frameworks}
  \label{table:comparison}
  \end{table}
\end{center}

\section{Source Code and Dependencies}

ELFI has been designed to be open source and modular, and can be extended through interfaces. For instance, it is possible to add new types of components, data stores or parallel clients. All the dependencies of ELFI are also open source.

ELFI is written in Python and is officially tested under Linux and MacOS but also works in Windows. The code style follows PEP 8 and documentation NumPy format. Code development uses the continuous integration practice with code review and automated tests to ensure the quality and usability of the software. The venue for distributing the source is GitHub that among the above features also allows anyone to raise issues regarding the software and make pull requests for new features.\footnote{The ELFI GitHub repository can be found at \url{https://github.com/elfi-dev/elfi}.} Online documentation is hosted in the Read The Docs.\footnotemark[1]
ELFI also has a community chat for the users.

\subsection*{Acknowledgements}
We would like to acknowledge support for this project from the Academy of Finland (Finnish Centre of Excellence in Computational Inference Research COIN) and grants
294238, 292334.
\\
We acknowledge the computational resources provided by the Aalto Science-IT project.

\vskip 0.2in
\bibliography{references}

\end{document}